\documentclass[10pt,twocolumn,letterpaper]{article}

\usepackage{cvpr}
\usepackage{times}
\usepackage{epsfig}
\usepackage{graphicx}
\usepackage{amsmath}
\usepackage{amssymb}
\usepackage{booktabs}
\usepackage{comment}
\usepackage{dsfont}
\usepackage{microtype}
\usepackage[ruled,linesnumbered,lined,commentsnumbered]{algorithm2e}
\usepackage[subrefformat=parens]{subcaption}

\usepackage[pagebackref=true,breaklinks=true,letterpaper=true,colorlinks,bookmarks=false]{hyperref}

\DeclareMathOperator*{\argmax}{arg\,max}

\makeatletter
\newcommand{\printfnsymbol}[1]{%
  \textsuperscript{\@fnsymbol{#1}}%
}

\makeatother

\cvprfinalcopy % *** Uncomment this line for the final submission
\def\cvprPaperID{8114} % *** Enter the CVPR Paper ID here

\begin{document}
\title{Two-stage Discriminative Re-ranking for Large-scale Landmark Retrieval}
\author{
~~~~~Shuhei Yokoo~~~~~\\
~~~~~University of Tsukuba~~~~~\\
~~~~~{\tt\small yokoo@cvlab.cs.tsukuba.ac.jp}~~~~~
\and
Kohei Ozaki~~~~~\\
Preferred Networks, Inc.\thanks{Work done while at Recruit Technologies Co.,Ltd.}~~~~~\\
{\tt\small ozaki@preferred.jp}~~~~~
\and
~~~~~~~~~~~~~~~~~Edgar Simo-Serra~~~~~~~~~~\\
~~~~~~~~~~~~~~~~~Waseda University~~~~~~~~~~\\
~~~~~~~~~~~~~~~~~{\tt\small ess@waseda.jp}~~~~~~~~~~
\and
~~~~~Satoshi Iizuka~~~~~\\
~~~~~University of Tsukuba~~~~~\\
~~~~~{\tt\small iizuka@cs.tsukuba.ac.jp}~~~~~
}
% \author[1]{Shuhei Yokoo}
% \author[2]{Kohei Ozaki}
% \author[3]{Edgar Simo-Serra}
% \author[1]{Satoshi Iizuka}
% \affil[1]{\normalsize University of Tsukuba \email{yokoo@cvlab.cs.tsukuba.ac.jp}}
% \affil[2]{\normalsize Preferred Networks, Inc.\thanks{Work done while at Recruit Technologies Co.,Ltd.}}
% \affil[3]{\normalsize Waseda University}
% \affil[]{\tt\small yokoo@cvlab.cs.tsukuba.ac.jp,ozaki@preferred.jp,ess@waseda.jp,iizuka@cs.tsukuba.ac.jp}
% \and \\
% {\normalsize Preferred Networks, Inc.\thanks{Work done while at Recruit Technologies Co.,Ltd.}}\\
% \and
% {\normalsize Waseda University}\\
% \and
% {\normalsize University of Tsukuba}\\
% \and
% {\tt\small ozaki@preferred.jp}
% {\tt\small ess@waseda.jp}
% {\tt\small iizuka@cs.tsukuba.ac.jp}
% {\tt\small yokoo@cvlab.cs.tsukuba.ac.jp}

% \author{
% Shuhei Yokoo\\
% {\normalsize University of Tsukuba}\\
% {\tt\scriptsize yokoo@cvlab.cs.tsukuba.ac.jp}
% \and
% Kohei Ozaki\\
% {\normalsize Preferred Networks, Inc.\thanks{Work done while at Recruit Technologies Co.,Ltd.}}\\
% {\tt\small ozaki@preferred.jp}
% \and
% Edgar Simo-Serra\\
% {\normalsize Waseda University}\\
% {\tt\small ess@waseda.jp}
% \and
% Satoshi Iizuka\\
% {\normalsize University of Tsukuba}\\
% {\tt\footnotesize iizuka@cs.tsukuba.ac.jp}
% }


\maketitle

\begin{abstract}
We propose an efficient pipeline for large-scale landmark image retrieval that
addresses the diversity of the dataset through two-stage discriminative re-ranking. Our
approach is based on embedding the images in a feature-space using a
convolutional neural network trained with a cosine softmax loss. Due to
the variance of the images, which include extreme viewpoint changes such as
having to retrieve images of the exterior of a landmark from images of the
interior, this is very challenging for approaches based exclusively on visual
similarity. Our proposed re-ranking approach improves the results
in two steps: in the sort-step, $k$-nearest neighbor search with soft-voting to sort
the retrieved results based on their label similarity to the query images, and
in the insert-step, we add additional samples from the dataset that
were not retrieved by image-similarity. This approach allows overcoming the low
visual diversity in retrieved images. In-depth experimental results show that
the proposed approach significantly outperforms existing approaches on the
challenging Google Landmarks Datasets.
Using our methods, we achieved 1st place in the Google Landmark Retrieval 2019 challenge and
3rd place in the Google Landmark Recognition 2019 challenge on Kaggle. Our code is publicly available here: \url{https://github.com/lyakaap/Landmark2019-1st-and-3rd-Place-Solution}
\end{abstract}

\begin{figure}[t]
\centering
  \includegraphics[width=\linewidth]{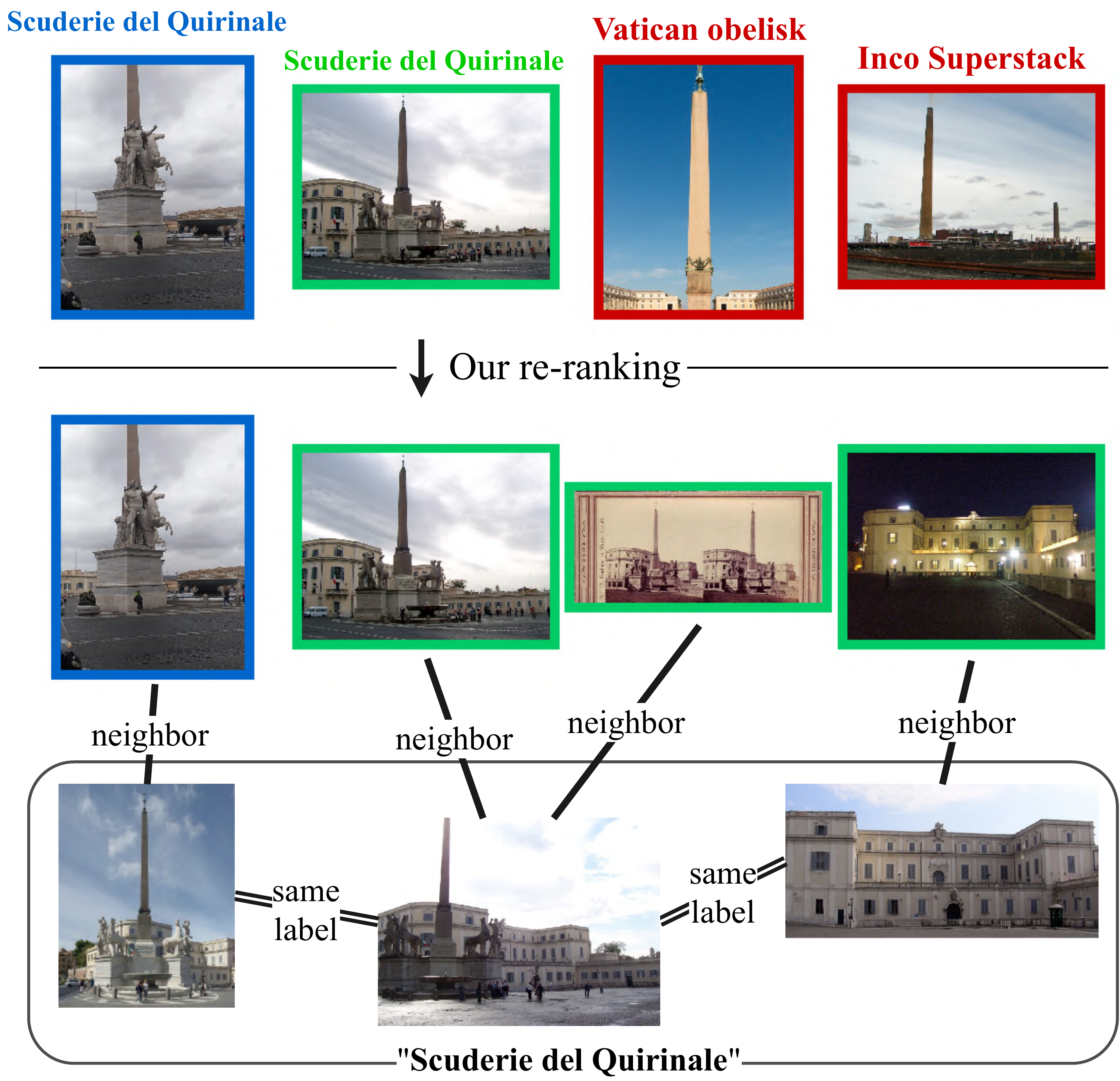}
  \caption{
  An example of improving the top-3 retrieved results from the Google Landmarks Dataset v2.1
  with our re-ranking approach.  The first row shows the result of a $k$-nearest neighbor ($k$-NN) search in the embedding space, the second row shows the result after the
  sort-step of our re-ranking, and the third row is the result after both the
  sort-step and insert-step of our re-ranking.  In each step, incorrect samples
  are replaced by correct samples based on their neighbors in label-space.
  Query images are in blue, correct samples are in green and incorrect samples
  are in red.
%   Best viewed in color.
  }
  \vspace{-5.0mm}
  \label{fig:teaser}
\end{figure}

\begin{figure*}[t]
\centering
  \includegraphics[width=0.9\linewidth]{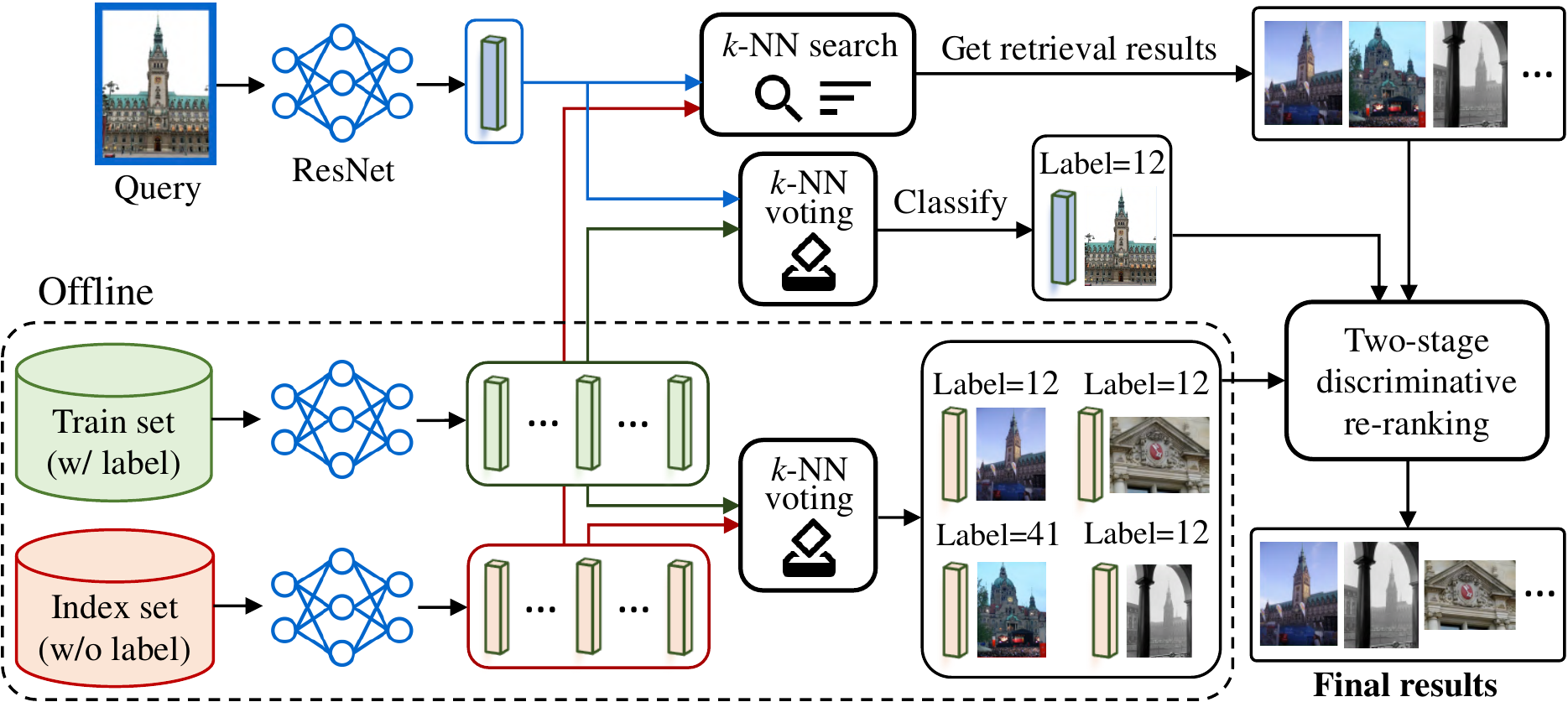}
  \caption{
%   Overview of our pipeline. Our model is based on ResNet-101 trained with train set of GLD-v1 and cleaned GLD-v2 using the ArcFace loss.
%   This yields compact image descriptors which can be used for image retrieval using $k$-NN search.
%   Finally, a two-stage discriminative re-ranking approach is used to refine and improve the retrieved images.
    Overview of our approach. In an offline step, image embeddings of the train set and index set are calculated, and the instance-id of each index set image is predicted by $k$-NN soft-voting.
    Afterward, the same prediction is performed for a query online, and initial retrieval results are obtained by $k$-NN search.
    For re-ranking, we assign ``positive'' or ``negative'' to the retrieval results based on their prediction results.
    Final results are obtained by two-stage discriminative re-ranking.
  }
  \label{fig:pipeline}
\end{figure*}

\vspace{-5mm}
\section{Introduction}
Image retrieval is a fundamental problem in computer vision where given a query image,
similar images must be found in a large dataset. In the case of landmark
images, the variation between points of view and different parts of the
landmark can be extreme, proving challenging for humans without deep knowledge
of the landmark in question. One such complicated example is shown in
Fig.~\ref{fig:teaser}. The Scuderie del Quirinale is very visually
similar to other structures such as the Vatican obelisk and the Inco
Superstack, leading to erroneous retrievals. Our proposed re-ranking approach
is able to exploit labeled information from the training dataset to improve the
retrieval results, even when the correct images are very visually dissimilar
such as drawings, different viewpoints, diverse illumination, etc.

Instance image retrieval can be seen as the task of converting the image
information into an embedding where similar images are nearby. Similar to
recent approaches, we focus on learning this embedding with a convolutional
neural network (CNN). We adopt a cosine softmax loss to train the neural
network for the retrieval task. Afterward, instead of simply using the
distance in the embedding space to find related images, we exploit the label
information to perform re-ranking.  Our re-ranking is based on a two-step
approach. In the \textit{sort-step}, a discriminative model based on $k$-NN
search with soft voting which allows us to sort the initial retrieved results
such that results more label-similar to the query image are given higher
priority. In the \textit{insert-step}, images that were originally not
retrieved are inserted into the retrieval results based on the same
discriminative model.  This combined approach shows a significant improvement
over existing approaches.

Noh \etal~\cite{iccvNohASWH17} has recently provided a
challenging dataset named Google Landmarks Dataset v1 (GLD-v1) for instance-level landmark image retrieval. For each
landmark, there is a diversity of images including both interior and exterior
images. Being able to identify the images without context is very challenging,
and in many cases, positive pairs have a very different visual appearance. More
recently, the dataset has been expanded in a second version (GLD-v2) to be more
complex and challenging. We focus on the retrieval task in this challenging
setting which due to being recent has not been fully explored yet.

Although the GLD-v2 dataset is a significant improvement over the previous
version, consistency and quality are still significant open issues that can be
very detrimental to results in the retrieval task. For this purpose, we also
propose an automatic data cleaning approached based on filtering the training
data. Although this reduces the dataset size and training budget, it ends up
being beneficial to overall performance of the model.

To summarize our contributions, (1) an effective pipeline for high quality
landmark retrieval, (2) a re-ranking approach based on exploiting label
information, (3) results that significantly outperform existing approaches on
challenging datasets.

\section{Related Work}

\noindent{\textbf{Instance Image Retrieval.}}
Image retrieval is usually posed as a problem of finding an image embedding in
which similar images have small distance, and has been traditionally done based
on local descriptor based
methods~\cite{SivicZ03,csurka2004BoVW,PerronninLSP10FisherVector,JegouPDSPS12VLAD}, including the popular 
SIFT~\cite{Lowe04SIFT}, RootSIFT~\cite{ArandjelovicZ12DQE}, and SURF~\cite{BayTG06SURF}.
% are classical hand-crafted local descriptors.
%These local descriptors are extracted by key-point detector such as Hessian-Affine detector~\cite{PerdochCM09HesAff}.
% RootSIFT is a improved version of SIFT by applying L1-normalization and element-wise square root so to SIFT.
Bag-of-Words~\cite{SivicZ03,csurka2004BoVW} model and its variants
(VLAD~\cite{JegouPDSPS12VLAD}, Fisher Vector~\cite{PerronninLSP10FisherVector})
have been popular in image retrieval previous to the advent of
learning-based approaches, and construct image embeddings by aggregating local
descriptors.  More recently, DELF~\cite{iccvNohASWH17} has been proposed as a deep learning-based local descriptor method, which uses the
attention map of CNN activation learned by only image-level annotation.  See
\cite{ZhengYT18IRSurvey} for a survey of instance image retrieval.

After the emergence of deep learning, many image retrieval methods based on deep
learning have been presented. Most recent image retrieval approaches are based
on deep
learning~\cite{RazavianASC14Offtheshelf,BabenkoL15SPoC,KalantidisMO16CroW,RMAC,NetVLAD,EndToEndDIR2017,Radenovic2018FinetuningCI}.
Both utilizing off-the-shelf CNN activations as an image
embedding~\cite{RazavianASC14Offtheshelf,BabenkoL15SPoC,KalantidisMO16CroW,RMAC}
and further fine-tuning to specific
datasets~\cite{NetVLAD,EndToEndDIR2017,Radenovic2018FinetuningCI} are popular
approaches.
An extension of VLAD called NetVLAD which is differentiable and trainable in an
end-to-end fashion 
%by replacing hard assignment with soft assignment
has also
been recently proposed~\cite{NetVLAD}.
% Triplet losses have also shown~\cite{WangSLRWPCW14TripletLoss1,HofferA14TripletLoss2} efficiency in learning image embeddings.
%, constructing image pair by GPS information.
Gordo \etal~\cite{EndToEndDIR2017} proposed using a region proposal network to
localize the landmark region and training a triplet network in an end-to-end
fashion.
%Radenovi{\'c} \etal~\cite{Radenovic2018FinetuningCI} proposes generalized mean pooling (GeM) and supervised-whitening.

The current state-of-the-art local descriptor based method is
D2R-R-ASMK~\cite{TeichmannAZS19D2R} along with spatial
verification~\cite{PhilbinCISZ07Oxford}.  D2R-R-ASMK is a regional aggregation
method comprising a region detector based on ASMK (Aggregated Selective Match
Kernels)~\cite{ToliasAJ16ASMK}. ASMK is one of the local feature aggregation
techniques.  The current state-of-the-art CNN global descriptor method is that
of Radenovi{\'c} \etal~\cite{Radenovic2018FinetuningCI} which employs an AP
loss~\cite{LandmarkListWiseLoss} along with re-ranking
methods~\cite{Yang2019EfficientIR,EGT2019}.
% Combination of local descriptor and global descriptor can achieve even higher accuracy complement each other~\cite{EGT2019}.
We construct our pipeline mainly based on latter strategy and show that by
using a two-stage discriminative re-ranking approach, we are able to obtain
results favorable to the existing approaches.

\noindent{\textbf{Retrieval Loss Functions.}}
Instance image retrieval requires image embedding that captures the similarity
well, and the loss used during learning plays an important role.
%Same instances should have similar embedding and different instances should have dissimilar embedding conversely.
%Learning metrics that satisfies such requirements is extremely important.
Using CNN off-the-shelf embeddings has been effective for image
retrieval~\cite{RazavianASC14Offtheshelf,BabenkoL15SPoC,KalantidisMO16CroW,RMAC}.
Babenko and Lempitsky~\cite{BabenkoL15SPoC} proposed using sum-pooling of CNN
activation, and Lin~\etal~\cite{RMAC} proposed max-pooling of multiple regions of
CNN activation.
However, training  specifically for the task of instance
retrieval has shown more effective with contrastive
loss~\cite{ChopraHL05ContrastiveLoss} and triplet
loss~\cite{WangSLRWPCW14TripletLoss1,HofferA14TripletLoss2,FaceNetTripletLoss15}
being some of the more used losses in image
retrieval~\cite{Radenovic-ECCV16,EndToEndDIR2017,Radenovic2018FinetuningCI,REMAP}.
%Contrastive loss and triplet loss both are initially proposed for face verification. 
Recently, the AP loss~\cite{LandmarkListWiseLoss}, which optimizes the global
mean average precision directly by leveraging list-wise loss formulations, has been proposed and achieved state-of-the-art results.
In face recognition field, recently cosine softmax losses~\cite{ranjan2017L2constrainedSoftmax,LiuWYY16SphereFace,LiuWYLRS17SphereFace,CosFace2018,WangCLL18CosFaceJournal,ArcFace2018,ZhangZQWL19Adacos}
have shown astonishing results and have become more favorable than other
losses~\cite{Masi0HN18FaceRecognitionSurvey}.
Cosine softmax losses impose L2-constraint to the features which
restricts them to lie on a hypersphere of a fixed radius, with popular
approaches being SphereFace~\cite{LiuWYY16SphereFace,LiuWYLRS17SphereFace},
ArcFace~\cite{ArcFace2018}, and
CosFace~\cite{CosFace2018,WangCLL18CosFaceJournal}, using  multiplicative angular
margin penalty, additive angular margin penalty, and additive cosine margin penalty, respectively.
While contrastive loss and triplet loss require training techniques such as
hard negative mining~\cite{FaceNetTripletLoss15,NetVLAD}, cosine softmax losses
do not and easy to implement and stable in training.  We show their successes
are not only in face recognition but also in instance image retrieval by
comparative experiments.

\noindent{\textbf{Re-ranking Methods.}}
Re-ranking is a essential approach to enhance the retrieval results on the image
embedding. Query expansion (QE)-based techniques are simple and popular ways
of re-ranking for improving recall of retrieval system.
AQE~\cite{ChumPSIZ07AQE} is the first work that applies query expansion in vision
field, and is based on averaging embeddings of top-ranked images retrieved by an initial
query, and using the averaged embedding as a new query.
%This new query is re-issued and retrieved results are enhanced.
$\alpha$QE~\cite{Radenovic2018FinetuningCI} uses weighted average of descriptors of top-ranked images. Heavier weights are put on as the rank gets higher.
DQE~\cite{ArandjelovicZ12DQE} uses an SVM classifier and its signed distance from the decision boundary for re-ranking.
Spatial verification (SP)~\cite{PhilbinCISZ07Oxford,PerdochCM09HesAff} is a
method that checks the geometric consistency using local descriptors and
RANSAC~\cite{Fischler:1981:RSC:358669.358692}, can be combined with QE to
filter images used for expansion~\cite{ChumPSIZ07AQE}. SP can be used as
re-ranking~\cite{ChumMPM11TotalRecall2,RITAC18} to improve precision, but it
has an efficiency problem.  Therefore, it is performed generally on a shortlist of
top-ranked images only.
HQE~\cite{ToliasJ14HQE} leverages Hamming Embedding~\cite{JegouDS08Hamming} to filter images instead of SP.

Diffusion~\cite{DonoserB13Diffusion} is major manifold-based approach, also
known as similarity propagation, which can also be used for re-ranking.  Many
diffusion approaches are proposed for enhancing the performance of instance
image
retrieval~\cite{DonoserB13Diffusion,IscenTAFC17,IscenATFC18,Yang2019EfficientIR}.
Diffusion can capture the image manifold in the feature space by random-walk on
$k$-NN graph.  Because diffusion process tends to be expensive, spectral
methods have been proposed to reduce computational cost~\cite{IscenATFC18}, and
Yang \etal~\cite{Yang2019EfficientIR} proposes decoupling diffusion into online
and offline processes to reduce online computation.  EGT~\cite{EGT2019} is
a recently proposed $k$-NN graph traversal algorithm, which outperforms diffusion
methods in terms of performance and efficiency.

Conventional re-ranking methods are unsupervised, which means they do not
consider label information even when label information is available. In
contrast, our re-ranking method can exploit label information, commonly
available in many problems, and shows excellent performance in landmark
retrieval tasks.

\section{Method}
\label{sec:method}

Our approach consists of training an embedding space using a cosine
softmax loss to train a CNN. Afterward,
retrieval is done based on $k$-NN search which is corrected and improved using
two-stage discriminative re-ranking.

\subsection{Embedding Model}

Our model is based on a CNN that embeds each image
into a feature-space amenable for $k$-NN search. Our model is based on a
ResNet-101~\cite{resnet} augmented with Generalized
Mean (GeM)-pooling~\cite{Radenovic2018FinetuningCI} to aggregate the spatial
information into a global descriptor.

The reduction of a descriptor dimension is crucial since it dramatically affects
the computational budget and alleviates the risk of over-fitting. We reduce the
dimension to 512 from 2048 by adding a fully-connected layer after the GeM-pooling layer.
Additionally, a one-dimensional Batch Normalization~\cite{bn} after the
fully-connected layer is used to improve the generalization ability.

Training is done using the ArcFace~\cite{ArcFace2018} loss with $L_2$ weight
regularization defined as follows
% we set s to 30, m to 0.3.
% we fix the individual weight |W| = 1 by l2 normalisation.
% we also fix the embedding feature |x| by l2 normalisation.
% m is a margin hyperparameter
% s is a scaling hyperparameter
% N is batch size
% y_i is target class (class of x_i)
% x_i is image embedding
% W_j denotes the j-th column of the weight W
% W_j \in \mathbb{R}^d, W \in \mathbb{R}^d \times n
\begin{align}
{L} &=-\frac{1}{N}\sum_{i=1}^{N}\log\frac{e^{s(\cos(\theta_{y_i}+m))}}{e^{s(\cos(\theta_{y_i}+m))}+\sum_{j=1,j\neq  y_i}^{n}e^{s\cos\theta_{j}}} \nonumber \\
%   &+ \beta (\sum_{i=1}^N \|W\|_2^2 + \| W_M \|_2^2) \;,
&+ \beta (\|W\|_2^2 + \| W_M \|_2^2) \;,
\end{align}
\noindent with
\begin{align}
\theta_{y_i} = W_{y_i}^\intercal \; f(x_i; W_M) \;\; \text{and} \;\;
\theta_{j} = W_{j}^\intercal \; f(x_i; W_M),
\end{align}
\noindent where $x_i$ is the input image with target class $y_i$, $N$ is the
batch size, $W$ denotes the weights of the last layer, $W_M$ is the parameters of the whole network excluding the last layer, $f(x;W_M)$ is the
embedding of $x$ using $W_M$, $s$ is a scaling hyperparameter, and
$m$ is a margin hyperparameter. We note that $\|W\|_2=1$ and $\|x_i\|_2=1$ is
enforced by normalizing at every iteration.

\subsection{Two-stage Discriminative Re-ranking}
The diversity of images belonging to the same instance is one of the main
problems in image retrieval.  For example, an instance of church may contain
diverse samples, such as outdoor and indoor images.  These images are extremely
hard to identify as the same landmark without any context.  Furthermore, the
visual dissimilarity makes it nearly impossible to retrieve them using only
visual-based embeddings.  To overcome this issue, we propose two-stage
discriminative re-ranking that exploits the label information.
An overview of our re-ranking approach is shown in Fig.~\ref{fig:pipeline}.

Our proposed method is composed of an auxiliary offline step and two re-ranking stages.
Suppose we have a query, an index set and a train set.
The index set is a database for which we perform image retrieval and has no labels, only images.
%% The train set is not a database for which we perform image retrieval and has labeled images.
First, we predict the instance-id of each sample from the index set by $k$-NN
search with soft-voting, where each sample from the index set is regarded as a
query, and the train set as a database.
%% database -> dictionary の方がわかりやすい？
%% instance-id -> label の方がわかりやすい？

The score of each instance-id is calculated by accumulating the cosine
similarities of the $k$ nearest samples as follows
\begin{align}
v(x,c) &= \frac{1}{k}\sum_{x'\in \mathcal{N}(x)} f(x')^\intercal \; f(x) \cdot \mathds{1}(\text{label}(x') = c), %\nonumber \\
%\hat{l}_{x} &= \argmax_{c} v(c), \ \hat{s}_x = \max_{c} v(c)\;,
\end{align}
\noindent where $f(\cdot) \in \mathbb{R}^d$ is the feature embedding function
with $W_M$ omitted for brevity, $\mathcal{N}(x)$ is the set of $k$ nearest
neighbours in the train set, and $\mathds{1}(\cdot)$ is an indicator
function. The prediction then becomes the class that maximizes $v(x,c)$ for
$x$. The index set prediction can be computed in an offline manner once.

When a query is given, its instance-id is also predicted in the same way described above.
Index set samples that are
predicted to be the same id of the query sample are treated as ``positive
samples'', and those of different id as ``negative samples'', and play an
important role in our re-ranking approach.

\begin{figure}[t]
\centering
\includegraphics[width=0.9\linewidth]{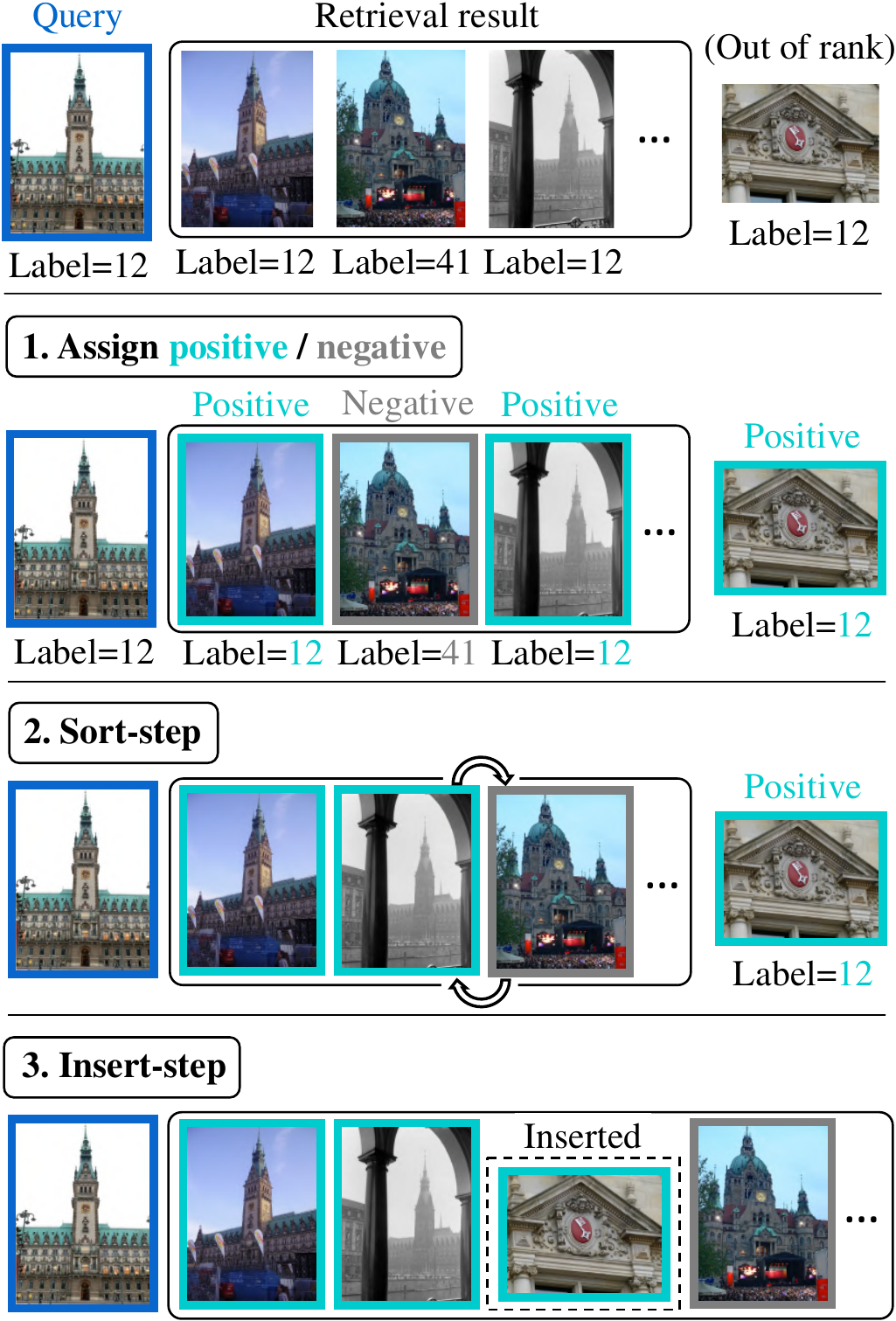}
\caption{
   Overview of our re-ranking procedure.
      ``Positive'' represents the samples predicted the same id as the predicted id of a query sample.
      ``Negative'' represents the samples predicted the different id from a predicted id of a query sample.
      Re-ranking is performed in each step based on their predicted id results.
      %   Query images are in blue, positive samples are in green and negative samples are in red.
}
\vspace{-4mm}
\label{fig:reranking}
\end{figure}

Our re-ranking method is illustrated in Figure~\ref{fig:reranking} and consists
of a \textit{sort-step} and \textit{insert-step}. The top row in the figure
shows a query (in blue) and retrieved samples from the index set by $k$-NN
search with positive samples shown in green and negative samples shown in red.
Here, we consider images on the left to be more relevant to the query than the
ones on the right. 
%Therefore, the right-most positive sample is considered less relevant than the negative sample on its left due to several factors (e.g., lighting, occlusion).
%It is desirable to ignore such trivial conditions for retrieval landmarks. 
In the \textit{sort-step}, positive samples are moved to the left of the
negative samples in the ranking, maintaining the relative order of them.  This
re-ranking step can make results more reliable, becoming less dependent on
factors such as lighting, occlusions, etc.

In the \textit{insert-step}, we insert positive samples from the entire index
set, which are not retrieved by the $k$-NN search, after the re-ranked positive
samples in descending order of scores which is calculated by $k$-NN cosine
similarities.  This step enables us to retrieve visually dissimilar samples to a
query by utilizing the label information of the train set.  Here, the
predicted instance-id may not always be reliable, especially when the
prediction score is low.  If the instance of query does not exist in the
train set, there is a tendency that the prediction score becomes very low.
Thus, we do not perform insertion to the sample of which the sum of the
prediction score between query and sample considered to be inserted is lower
than a threshold $\tau_\mathrm{score}$ to deal with such a situation.

\noindent{\bf Discussion.}
Our re-ranking method can be applied when a train set exists and there are some
overlaps of instances from the train set and a index set (database).
Although conventional instance image retrieval datasets have no instance overlap between the train set and the index set
to measure generalization performance of methods, it is natural to have instance overlap between them
in most real situations.  For example, in a potential landmark image search
system, some users may upload their landmark photos with a landmark name, which
can be a label.  Thus, using these meta information is natural and essential to
improve search results.

In the evaluation on the GLD-v2 dataset, the train set is not used.  Since our re-ranking
follows with the GLD-v2 evaluation criteria, we constructed the algorithm
considering that the samples from the train set are not a target of retrieval.
However, when considering the actual retrieval system, the train set can also
be considered to be part of the database to be searched.  Even in such cases,
our re-ranking can be naturally expanded.  Specifically, in our re-ranking, it
is necessary to predict the instance-id of each sample of the index set in
advance. However, the instance-ids of samples from the train set are known.
Thus, we can use these instance-ids of train set samples by setting the prediction score 1.0.  By doing so, our re-ranking can be
executed without changing in other steps, no matter whether retrieved samples are from
the index set or the train set.

\begin{table}[t]
\centering
\begin{tabular}{l r r}
\toprule
Dataset (train set) & \# Samples & \# Labels \\
\midrule
GLD-v1             & 1,225,029  & 14,951    \\
GLD-v2/2.1         & 4,132,914  & 203,094   \\
GLD-v2/2.1 (clean) & 1,580,470  & 81,313    \\
\bottomrule
\end{tabular}
\caption{Dataset statistics used in our experiments. The index and test images
are not included here. GLD-v2 and GLD-v2.1 only differ in the index set and test set and
thus are shown together for the train set.}
\label{tab:dataset_chara}
\vspace{-5mm}
\end{table}

\section{Dataset}
\label{sec:dataset}
The Google Landmarks Dataset (GLD) is the largest dataset of instance image
retrieval, which contains photos of landmarks from all over the world.  The
photos include a lot of variations, e.g., occlusion, lighting changes.  GLD has
three versions: v1, v2, and v2.1 and we overview their differences in
Table~\ref{tab:dataset_chara}.  GLD-v1~\cite{iccvNohASWH17} which is the first
version of GLD has released in 2018.  This dataset has more than 1 million
samples and around 15 thousand labels.  
GLD-v1 was created based on the algorithm described in~\cite{ZhengZSABBBCN09}, and uses visual features and GPS coordinates for ground-truth correction.  Simultaneously, the
Google Landmarks Challenge 2018 was launched and GLD-v1 was used at this
challenge.  Currently, we can still download the dataset, but cannot evaluate
with it since ground-truth was not released.
GLD-v2~\footnote{https://github.com/cvdfoundation/google-landmark}, used for the
Google Landmarks Challenge 2019, is
the largest worldwide landmark recognition dataset available at the time.  This
dataset includes over 5 million images of more than 200 thousands of different
landmarks.  It is divided into three sets: train, test, and index. Only samples
from the train set are labeled.
% The retrieval track asks us to find an image of the same instance (landmark) from the index set, while the recognition track asks us to answer the corresponding label defined in the train set.

%GLD-v1 was released after an automated data cleaning step, while GLD-v2 is the raw data.  
Since GLD-v2 was constructed by mining web landmark images without
any cleaning step, each category may contain quite diverse samples: for
example, images from a museum may contain outdoor images showing the building
and indoor images depicting a statue located in the museum. In comparison with the GLD-v1, there is significantly more noise in the annotations.
The GLD-v2.1 is a minor update of GLD-v2. Only ground truth of test set and index
set are updated.

\noindent{\bf Automated Data Cleaning.}
The train set of GLD-v2 is very noisy because it was constructed by mining web
landmark images without any cleaning step.  Furthermore, training with the
entire train set of GLD-v2 is complicated due to its huge scale.  Therefore, we
consider to automatically remove noises such as mis-annotation inspired
by~\cite{EndToEndDIR2017},  leading to reduction of dataset size and training
budget, while avoiding  adverse effects of the noise for deep metric learning.

To build a clean train set, we apply spatial
verification~\cite{PhilbinCISZ07Oxford} to filtered images by $k$-NN search.
% 具体的手順,
% 1. train set の画像 x_i について、top-k 近傍を取り出す
% 2. top-k 近傍のうち、同じlandmark_idの画像の高々100件に対して spatial verification を適用する。閾値 t_verify 以上の画像を verified images とする。
% 3. verified images が t_frequency 以上ある場合、x_i を cleaned dataset に追加する。
% (パラメータと更に詳細) t_frequency=2, t_verify=30. affine+DELF.
Specifically, cleaning the train set consists of a three-step process.
First, for each image descriptor $x_i$ in the train set, we search its 1000
nearest neighbors from the train set.  This image descriptor is obtained by our
embedding model learned from the GLD-v1 dataset.  Second, spatial verification is
performed on up to the 100 nearest neighbors assigned to the same label as
$x_i$.  For spatial verification, we use
RANSAC~\cite{Fischler:1981:RSC:358669.358692} with affine transformation and
deep local attentive features (DELF)~\cite{iccvNohASWH17}.  If an
inlier-count between $x_i$ and nearest neighbor image descriptor is greater than
30, we consider the nearest neighbor as a verified image.  Finally, if the count
of verified images in the second step reaches the threshold $\tau_{\text{freq}}$,
$x_i$ is added to the cleaned dataset.  We set $\tau_{\text{freq}} = 3$ in our
experiment.

This automated data cleaning is very costly due to the use of spatial verification,
however, it only has to be run once.  Table~\ref{tab:dataset_chara} summarizes
the statistics of the dataset used in our experiments.  We show the
effectiveness of using our cleaned dataset through our experiments in the
following sections.

% Our cleaned dataset is publicly available.
% https://www.kaggle.com/confirm/cleaned-subsets-of-google-landmarks-v2

\begin{table*}[t]
\centering
\begin{tabular}{lccccccccc}
\toprule
 & \multicolumn{2}{c}{GLD-v2} & \multicolumn{6}{c}{GLD-v2.1} \\ 
 \cmidrule(rl){2-3} \cmidrule(rl){4-9}
 & Private & Public & \multicolumn{3}{c}{Private} & \multicolumn{3}{c}{Public} \\ 
\cmidrule(rl){2-2} \cmidrule(rl){3-3} \cmidrule(rl){4-6} \cmidrule(rl){7-9}
Method   & {\small mAP@100} & {\small mAP@100} & {\small mAP@100} & {\small P@10} & {\small MeanPos} & {\small mAP@100} & {\small P@10} & {\small MeanPos} \\ 
\midrule
$k$-NN search             & 30.22 &	27.81 & 29.63 & 30.76 & 27.02 & 27.66 & 28.87 & 32.60 \\ 
\midrule
SP~\cite{RITAC18} & 23.75 & 22.40 & 23.29 & 24.72 & 28.72 & 22.15 & 23.46 & 33.18 \\
AQE~\cite{ChumPSIZ07AQE} & 32.17 & 30.47 & 31.60 & 32.97 & 27.44 & 30.28 & 31.35 & 31.54 \\
$\alpha$QE~\cite{Radenovic2018FinetuningCI} & 32.21 & 30.34 & 31.71 & 33.04 & 26.67 & 30.23 & 31.03 & 31.13 \\ 
Iscen \etal's DFS~\cite{IscenTAFC17} & 32.01 & 30.55 & 31.91 & 32.51 & 29.52 & 30.81 & 30.50 & 33.23 \\
Yang \etal's DFS~\cite{Yang2019EfficientIR} & 31.20 & 29.36 & 30.90 & 31.48 & 29.87 & 29.29 & 29.63 & 33.83 \\ 
EGT~\cite{EGT2019} & 30.33 & 28.44 & 31.00 & 32.89 & 34.82 & 29.77 & 30.74 & 38.19 \\ 
\midrule
Ours               & 36.85 & 34.89 & 36.04 & 36.27 & \textbf{24.43} & 34.41 & 33.40 & 29.23 \\ 
Ours + $\alpha$QE  & \textbf{37.34} & \textbf{35.59} & \textbf{36.55} & \textbf{36.68} & 24.44 & \textbf{35.12} & \textbf{33.85} & \textbf{28.11} \\ 
\bottomrule
\end{tabular}
\caption{
Comparison of our re-ranking against the other state-of-the-art re-ranking methods on top of our baseline.
We report mAP@100 in GLD-v2 and mAP@100, P@10, and MeanPos in GLD-v2.1.
mAP@100 is mean average precision at rank 100.
P@10 is mean precision at rank 10 and higher is better.
MeanPos is the mean position of the first relevant image (if no relevant image in top-100, use 101 as position) and lower is better.
}
\label{tab:res_reranking}
\vspace{-0.5mm}
\end{table*}

\begin{figure}[t]
\centering
  \includegraphics[width=0.9\linewidth]{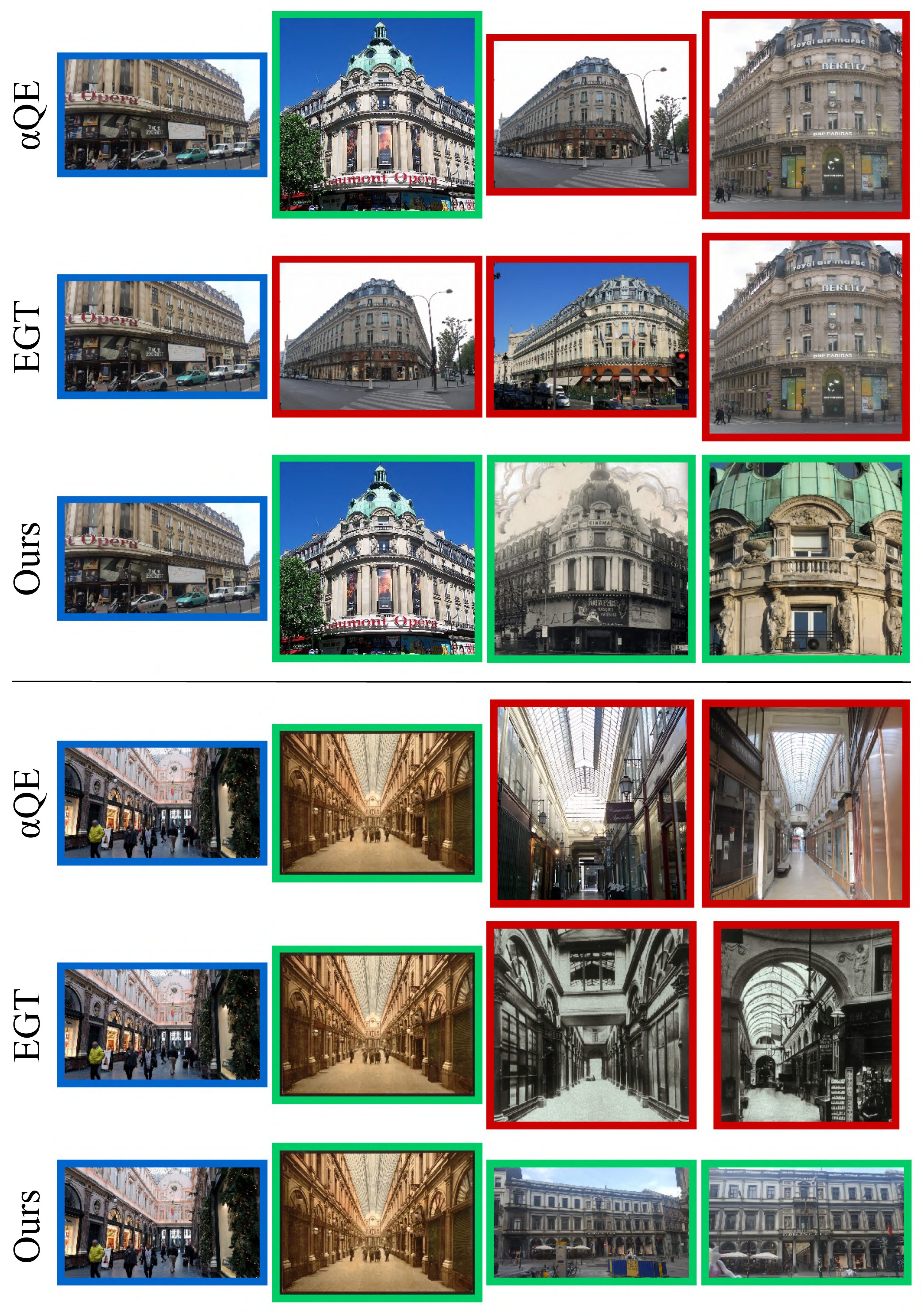}
  \vspace{-2.6mm}
  \caption{
%   Two examples of top-3 retrieved results from GLD-v2.1.  The first row is the
%   result of $\alpha$QE, and the second row is the result of EGT, and the third
%   row is the result of our approach.  Query images are in blue, correct samples
%   are in green and incorrect samples are in red.  Best viewed in color.
  Two examples of top-3 retrieved results from GLD-v2.1 using $\alpha$QE, EGT, and our approach. Query images are in blue, correct samples
  are in green and incorrect samples are in red.  Best viewed in color.
  }
  \label{fig:vis_compare}
  \vspace{-3.0mm}
\end{figure}

\begin{figure}[t]
\centering
  \includegraphics[width=0.9\linewidth]{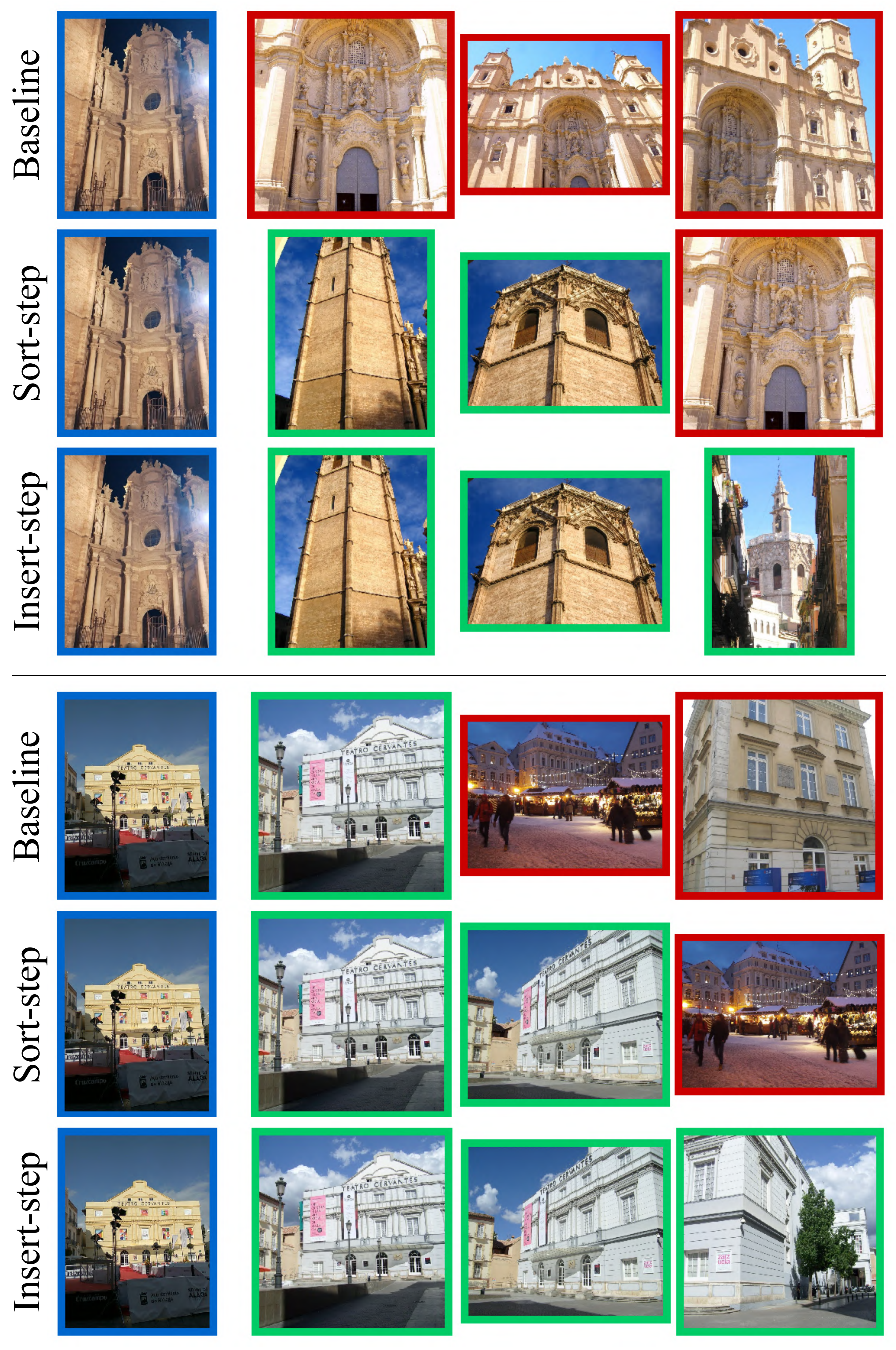}
  \vspace{-2.6mm}
  \caption{
  Two examples of top-3 retrieved results from GLD-v2.1, improved by using our
  re-ranking.  The first row is the result of $k$-NN search, and the second row
  is the result after the sort-step, and the third row is the result after the
  insert-step, including the sort-step. Query images are in blue, correct
  samples are in green and incorrect samples are in red.  Best viewed in color.
  }
  \label{fig:vis}
  \vspace{-3.0mm}
\end{figure}

\section{Experiments}

\subsection{Implementation Details}
\label{sec:impl}
%To obtain landmark image representation, convolutional neural networks are employed through our pipeline.
We pre-train the model on ImageNet~\cite{imagenet} and the train set of
GLD-v1~\cite{iccvNohASWH17} first, before being trained on cleaned GLD-v2 train
set with a cosine softmax loss.  We use $p=3.0$ for the Generalized
Mean-pooling, and use 512-dimension embedding space. We use a margin of 0.3 for
the ArcFace loss and $\beta=10^{-5}$ for the regularization term. For
re-ranking we use $\tau_\mathrm{score}=0.6$ and $k=3$ for $k$-NN soft-voting.

%In particular work, we use both the ArcFace~\cite{ArcFace2018} and CosFace~\cite{CosFace2018} losses with a margin of 0.3.

We train each network for 5 epochs with commonly used data augmentation methods
such as brightness shift, random cropping, and scaling. In particular, images
are randomly scaled between 80\% and 120\% of their original size and then
either cropping or zero-padding is used to return the image to the original
resolution, depending on whether the image was downscaled or upscaled.
Brightness is randomly modified by 0\% to 10\%.
When constructing mini-batches for training, the images are resized to be
the same size for efficient training.  This might cause distortions to the
input images, degrading the accuracy of the
network~\cite{BestPracticeInstanceRetrieval}. To avoid this, we choose
mini-batch samples so that they have similar aspect ratios, and resize them to
a particular size.  The size is determined by selecting tuple of width and
height from $[\;(512, 352), (512, 384), (448, 448), (384, 512), (352, 512)\;]$
depending on their aspect ratio.
% Revisitの評価時はoriginal resolution. GLD-v2評価時は [(800, 544), (800, 608), (800, 800), (608, 800), (544, 800)] の中から近いものを選ぶようにした（ミニバッチ推論したかったため）。

%Our implementation is based on PyTorch~\cite{Pytorch}, and four NVIDIA Tesla-V100 GPUs are used for training.
Model training is done by using the stochastic gradient descent
with momentum, where initial learning rate, momentum, and batch size are set to 0.001, 0.9, and 32, respectively.
The cosine annealing~\cite{SGDR2017CosineAnnealing} learning rate scheduler is used during training.

For other approaches we compare to, we follow the settings described in their
respective papers.  However, we have changed some hyperparameters which would
found to give non-competitive results.  In particular, spatial verification
(SP) follows the procedure from~\cite{RITAC18} except for using
DELF~\cite{iccvNohASWH17} trained with GLD-v1 as the local descriptor.  In
AQE~\cite{ChumPSIZ07AQE} and $\alpha$QE~\cite{Radenovic2018FinetuningCI}, the
number of retrieved results used for query expansion are set to 10 including
the query itself. The $\alpha$ of $\alpha$QE is set to 3.0.  SP is not used to
filter samples for the construction of a new query in QE different
from~\cite{ChumPSIZ07AQE}.  For Iscen \etal's diffusion
(DFS)~\cite{IscenTAFC17} and Yang \etal's diffusion
(DFS)~\cite{Yang2019EfficientIR}, the default hyperparameters are used.  The
threshold $t$ of EGT~\cite{EGT2019} is set to $inf$.
Hyperparameters of each method are tuned using the GLD-v2 Public split.

We use multi-scale feature extraction described in~\cite{EndToEndDIR2017} during test time in whole experiments. The resulting features are finally averaged and re-normalized.

\subsection{Evaluation Protocol}
We use the Google Landmarks Dataset (GLD)~\cite{iccvNohASWH17}, $\mathcal{R}$Oxford-5K~\cite{RITAC18}, and $\mathcal{R}$Paris-6K~\cite{RITAC18} for experiments.
GLD-v1 and GLD-v2 have three data splits: train, index and test set.
The train set of GLD-v1 and GLD-v2 is used for training.
Additionally, the train set of GLD-v2 is used as a train set for re-ranking.
The index set and the test set of GLD-v2 and GLD-v2.1 are used for our evaluation.
%% The samples from the test set are treated as queries.
The index set and the test set of GLD-v1 are not used for our evaluation since we cannot obtain ground-truth of GLD-v1 and use evaluation server.
Note that evaluation on GLD-v2 are performed on evaluation server of the competition page~\footnote{https://www.kaggle.com/c/landmark-retrieval-2019/submit} and it shows only mAP@100.
We report two split results, ``Private'' and ``Public''.
The Private split accounts for 67\% and the Public split accounts for 33\% of GLD-v2 and GLD-v2.1 respectively.

Additionally, $\mathcal{R}$Oxford-5K~\cite{RITAC18}, and
$\mathcal{R}$Paris-6K~\cite{RITAC18} are also used for the evaluation of loss
functions and dataset comparison.  $\mathcal{R}$Oxford-5K~\cite{RITAC18}, and
$\mathcal{R}$Paris-6K~\cite{RITAC18} are the revisited version of
Oxford~\cite{PhilbinCISZ07Oxford} and Paris~\cite{PhilbinCISZ08Paris}.  We
follow the Hard evaluation protocol~\cite{RITAC18}.

\subsection{Comparison with Other Re-ranking Methods}

We evaluate our re-ranking method and other state-of-the-art re-ranking methods
on top of our baseline in Table~\ref{tab:res_reranking}, evaluating on the
GLD-v2 and GLD-v2.1 datasets.  Baseline is the retrieved results by $k$-NN
search using descriptors extracted by our trained model.  Surprisingly, spatial
verification (SP)~\cite{RITAC18} harms the performance drastically in contrast to the common
sense of instance image retrieval.
%, although this phenomenon has also reported in~\cite{TeichmannAZS19D2R}.
% quotation: "SP further boosts performance by about 3% mAP on ROxf. Surprisingly, it actually degrades performance on the RPar dataset, by about 2%"
After visual inspection of the results of SP, we hypothesize that this is
likely caused by a large number of instances that are very similar.
% ess: 下の文章が分かりません
% yokoo: RANSACによるinlier countsの大きさが
% 負例だけど局所的なパーツが一致している画像 ＞ 正例だけど撮影角度などが大きく違う画像
% という大小関係になってしまうことでスコアが下がってしまう事例が多くあるということが言いたかったです。具体的な例はhangoutのチャット欄に画像を貼りました。
There are many cases where the RANSAC inlier count increases artificially due
to geometrical consistency of partial region between even different instances,
degrading accuracy as a result.
% We hypothesize that this phenomenon might be induced by the unreliability of
% the inlier count of RANSAC randomness~\cite{Fischler:1981:RSC:358669.358692}.

Experimental results show that our approach outperforms the previous re-ranking
approaches on the challenging GLD dataset. Furthermore, a combination of ours
and $\alpha$QE boosts the performance, and it suggests that our re-ranking
method can be combined with existing re-ranking methods to further improve performance.
% Our re-ranking with ensemble 6 models gave the 1st place result in Google Landmark Challenge 2019 in Retrieval Track out of more than 100 teams.
A qualitative comparison with other approaches is shown in Fig.~\ref{fig:vis_compare}.
% Figure~\ref{fig:vis_compare} is qualitative comparisons among $\alpha$QE (top), EGT (middle) and ours (bottom).
We can see that our re-ranking can retrieve samples that have no visual clue to
query. These samples are failed to be retrieved with $\alpha$QE and EGT.

\subsection{Ablation Study}
We perform an ablation study and report the result in Table~\ref{tab:ablation}
to validate each step in our re-ranking approach.  We can see that both the
sort-step and insert-step significantly improve results with respect to the
$k$-NN search-only baseline.
%It illustrates each step improve the baseline, which is retrieved results by $k$-NN search only, in every metrics by a large margin.
%It can be seen that both of steps play important roles in our re-ranking procedure.

% \begin{table*}[t]
% \centering
% \begin{tabular}{lccccccccc}
% \toprule
%  & \multicolumn{2}{c}{GLD-v2} & \multicolumn{6}{c}{GLD-v2.1} \\ 
%  \cmidrule(rl){2-3} \cmidrule(rl){4-9}
%  & Private & Public & \multicolumn{3}{c}{Private} & \multicolumn{3}{c}{Public} \\ 
% \cmidrule(rl){2-2} \cmidrule(rl){3-3} \cmidrule(rl){4-6} \cmidrule(rl){7-9}
% Description   & mAP@100 & mAP@100 & mAP@100 & P@10 & MeanPos & mAP@100 & P@10 & MeanPos \\ 
% \midrule
% Baseline      & 30.22 &	27.81 & 29.63 & 30.76 & 27.02 & 27.66 & 28.87 & 32.60 \\
% + Sort-step	  & 33.79 &	30.91 & 33.18 & 34.45 & 26.20 & 30.69 & 31.64 & 32.25 \\ 
% + Insert-step &	\textbf{36.85} & \textbf{34.89} & \textbf{36.04} & \textbf{36.27} & \textbf{24.43} & \textbf{34.41} & \textbf{33.40} & \textbf{29.23} \\ 
% \bottomrule
% \end{tabular}
% \caption{Ablation study of each step of our re-ranking. We show the effect of
% adding the sort-step and both the sort-step and insert-step with respect to our
% strong baseline.}
% \label{tab:ablation}
% \end{table*}

\begin{table}[t]
\centering
\begin{tabular}{lccccccccc}
\toprule
Description   & Private & Public \\
\midrule
Baseline      & 30.22 &	27.81  \\
+ Sort-step	  & 33.79 &	30.91 \\ 
+ Insert-step &	\textbf{36.85} & \textbf{34.89} \\ 
\bottomrule
\end{tabular}
\caption{Ablation study of each step of our re-ranking. We show the effect of
adding the sort-step and both the sort-step and insert-step with respect to our
strong baseline on the GLD-v2 dataset.}
\label{tab:ablation}
%\vspace{-2.0mm}
\end{table}

Additionally, we show the top-3 ranked results of each step in
Fig.~\ref{fig:vis}. We can see that the baseline of $k$-NN search retrieves visually
similar images no matter if it shows the same landmark as the query image or not.
After each step, the correct images not retrieved as top rank samples due to the
visual dissimilarity are more emphasized and ranked higher.

We test the effect of hyperparameter $k$ of $k$-NN and $\tau_\mathrm{score}$ and 
report results in Table~\ref{tab:ablation_k}.
% In this experiments, we set prediction score threshold to 0(equivalent to perform insert-step even if the prediction score is low).
We can see that our re-ranking approach is fairly insensitive to the setting of
the hyperparameters.

\begin{table}[t]
\centering
\begin{tabular}{lcccccccccc}
\toprule
$k$   & $\tau_\mathrm{score}$ & Private & Public \\ 
\midrule
1 & 0.0 & 35.35 & 33.28 \\
1 & 0.6 & 35.35 & 33.28 \\
1 & 1.2 & 35.36 & 33.28 \\
3 & 0.0 & 36.77 & 34.88 \\
3 & 0.6 & \textbf{36.85} & \textbf{34.89} \\
3 & 1.2 & 35.76 & 33.12 \\
5 & 0.0 & 35.78 & 33.98 \\
5 & 0.6 & 35.88 & 34.12 \\
5 & 1.2 & 34.68 & 32.09 \\
\bottomrule
\end{tabular}
\caption{Effect of two-stage discriminative re-ranking hyperparameters $k$ used for
$k$-NN search and insert threshold $\tau_\mathrm{score}$ on the GLD-v2
dataset.}
\label{tab:ablation_k}
\vspace{-1.0mm}
\end{table}

\subsection{Comparison of Loss Functions}
% In this section, we compare our model with other state-of-the-art global descriptor models.
Table~\ref{tab:res_lossfun} shows the comparison results among loss functions when trained with GLD-v1.
ResNet-101~\cite{resnet} is used as backbone network in all loss function experiments.
In the triplet loss and AP loss, we use an implementation described
in~\cite{Radenovic2018FinetuningCI}, and~\cite{Radenovic2018FinetuningCI}
offers a state-of-the-art global descriptor model.  In CosFace~\cite{CosFace2018}
and ArcFace~\cite{ArcFace2018}, we use a model described in
Section~\ref{sec:impl} with a margin of 0.3. Note that we do not use supervised whitening in
CosFace and ArcFace experiments for the sake of simplicity. We set the
dimension of the global descriptor to 2048 in triplet loss and AP loss following
the setting of~\cite{Radenovic2018FinetuningCI,LandmarkListWiseLoss},
and 512 in CosFace and ArcFace.

\begin{table}[t]
\centering
\begin{tabular}{lcccc}
\toprule
Loss    & {\small Private}        & {\small Public}   & {\small $\mathcal{R}$Oxf}       & {\small $\mathcal{R}$Par}     \\ 
\midrule
% Softmax & 17.95          & 15.00     & 66.73          & 43.5           & 79.25          & 58.58          \\ 
TripletLoss~\cite{Radenovic2018FinetuningCI} & 18.94 & 17.14 & 43.61 & 61.39 \\
AP Loss~\cite{LandmarkListWiseLoss} & 18.71 & 16.30 & 40.87 & 61.62 \\
CosFace~\cite{CosFace2018} & \textbf{21.35} & \textbf{18.41}     & 44.78          & 62.95          \\ 
ArcFace~\cite{ArcFace2018} & 20.74          & 18.13          & \textbf{46.25} & \textbf{66.62} \\ 
\bottomrule
\end{tabular}
\caption{Comparison of loss functions.
We train with GLD-v1 and use ResNet-101~\cite{resnet} for the loss function comparison.
We report mAP@100 on GLD-v2 (Private and Public splits) and mAP on Hard evaluation protocol of $\mathcal{R}$Oxford-5K~\cite{RITAC18} ($\mathcal{R}$Oxf) and $\mathcal{R}$Paris-6K~\cite{RITAC18} ($\mathcal{R}$Par).
}
\label{tab:res_lossfun}
%\vspace{-2.0mm}
\end{table}

\begin{table}[t]
\centering
\begin{tabular}{lcccc}
\toprule
Dataset (GLD)              & Private        & Public         & $\mathcal{R}$Oxf       & $\mathcal{R}$Par \\
\midrule
v1                & 20.74          & 18.13                    & 46.25           & 66.62          \\ 
v2                & 27.81          & 24.97          	& 54.81 &	74.40 \\ 
v2-clean          & 28.83          & 26.86          	& 58.94 & \textbf{78.13} \\
v1 + v2       & 29.20          & 26.84                   & 56.59           & 77.35          \\ 
v1 + v2-clean & \textbf{30.22} & \textbf{27.81}  & \textbf{59.93}          & 77.82 \\ 
\bottomrule
\end{tabular}
\caption{Evaluation of the effect of the training dataset. We use the ArcFace loss with a ResNet-101 model for this experiment. We report mAP@100 on GLD-v2 (Private and Public splits) and mAP on Hard evaluation protocol of $\mathcal{R}$Oxford-5K~\cite{RITAC18} ($\mathcal{R}$Oxf) and $\mathcal{R}$Paris-6K~\cite{RITAC18} ($\mathcal{R}$Par).}
\label{tab:res_dataset}
\vspace{-1.0mm}
\end{table}

Although it is hard to compare the loss functions fairly due to the
implementation differences, CosFace and
ArcFace seem to outperform triplet loss and AP loss in multiple
benchmarks. CosFace outperforms to ArcFace in Private and Public set of GLD-v2.  ArcFace outperforms to CosFace in the other metrics.

\subsection{Datasets}
We perform experiments to validate the influence of the training dataset.
Table~\ref{tab:res_dataset} shows the results of comparison with various dataset combination.
``v1'' denotes the train set of GLD-v1, and ``v2'' denotes the train set of GLD-v2.
``v2-clean'' is the GLD-v2 train set cleaned by the automated way described in Section~\ref{sec:dataset}.
We find that training with v2 significantly increases performance with respect to v1.
%Regarding the difference between GLD-v1 and GLD-v2, training with GLD-v2 overwhelmingly surpass the performance to the GLD-v1.
The result using v2-clean for training outperforms the result using v2 either with and without v1 pre-training, in spite of reducing the sample size by three. Using v2-clean with v1 pre-training gives the best results overall.
%We adopt model trained with ``GLD-v1 + GLD-v2-clean'' for other experiments.
%ArcFace is used for loss function and ResNet-101 is used for backbone through all dataset experiments.

% \begin{table*}[t]
% \centering
% \begin{tabular}{lcccccc}
% \toprule
% Dataset               & Private        & Public \\ 
% \midrule
% GLD-v1                & 20.74          & 18.13 \\ 
% GLD-v2                & 27.81          & 24.97 \\
% GLD-v2-clean          & 28.83          & 26.86 \\
% GLD-v1 + GLD-v2       & 29.20          & 26.84 \\ 
% GLD-v1 + GLD-v2-clean & \textbf{30.22} & \textbf{27.81} \\ 
% \bottomrule
% \end{tabular}
% \caption{Evaluation of the effect of the training dataset. We report mAP@100 on GLD-v2 Private and Public splits. We use the ArcFace loss with a ResNet-101 model for this experiment.}
% \label{tab:res_dataset}
% \end{table*}

% \section{Conclusion and Discussion}
\section{Conclusion}

We have presented an efficient pipeline for retrieval of landmark images from
large datasets. Our work leverages recent approaches and we propose a
discriminative two-step re-ranking method that shows significant improvements with
respect to existing approaches. In-depth experimental results corroborate the
efficacy of our approach.

\clearpage
{\small
\bibliographystyle{ieee_fullname}
\bibliography{egbib}

\begin{thebibliography}{10}\itemsep=-1pt

\bibitem{NetVLAD}
Relja Arandjelovic, Petr Gron{\'{a}}t, Akihiko Torii, Tom{\'{a}}s Pajdla, and
  Josef Sivic.
\newblock Netvlad: {CNN} architecture for weakly supervised place recognition.
\newblock {\em TPAMI}, 40(6):1437--1451, 2018.

\bibitem{ArandjelovicZ12DQE}
Relja Arandjelovic and Andrew Zisserman.
\newblock Three things everyone should know to improve object retrieval.
\newblock In {\em CVPR}, pages 2911--2918, 2012.

\bibitem{BabenkoL15SPoC}
Artem Babenko and Victor~S. Lempitsky.
\newblock Aggregating local deep features for image retrieval.
\newblock In {\em ICCV}, pages 1269--1277, 2015.

\bibitem{BayTG06SURF}
Herbert Bay, Tinne Tuytelaars, and Luc~Van Gool.
\newblock {SURF:} speeded up robust features.
\newblock In {\em ECCV}, pages 404--417, 2006.

\bibitem{EGT2019}
Cheng Chang, Guangwei Yu, Chundi Liu, and Maksims Volkovs.
\newblock Explore-exploit graph traversal for image retrieval.
\newblock In {\em CVPR}, 2019.

\bibitem{ChopraHL05ContrastiveLoss}
Sumit Chopra, Raia Hadsell, and Yann LeCun.
\newblock Learning a similarity metric discriminatively, with application to
  face verification.
\newblock In {\em CVPR}, pages 539--546, 2005.

\bibitem{ChumMPM11TotalRecall2}
Ondrej Chum, Andrej Mikul{\'{\i}}k, Michal Perdoch, and Jiri Matas.
\newblock Total recall {II:} query expansion revisited.
\newblock In {\em CVPR}, pages 889--896, 2011.

\bibitem{ChumPSIZ07AQE}
Ondrej Chum, James Philbin, Josef Sivic, Michael Isard, and Andrew Zisserman.
\newblock Total recall: Automatic query expansion with a generative feature
  model for object retrieval.
\newblock In {\em ICCV}, pages 1--8, 2007.

\bibitem{csurka2004BoVW}
Gabriella Csurka, Christopher Dance, Lixin Fan, Jutta Willamowski, and
  C{\'e}dric Bray.
\newblock Visual categorization with bags of keypoints.
\newblock In {\em ECCVW}, 2004.

\bibitem{imagenet}
Jia Deng, Wei Dong, Richard Socher, Li-Jia Li, Kai Li, and Li Fei-Fei.
\newblock Imagenet: A large-scale hierarchical image database.
\newblock In {\em CVPR}, pages 248--255, 2009.

\bibitem{ArcFace2018}
Jiankang Deng, Jia Guo, Niannan Xue, and Stefanos Zafeiriou.
\newblock Arcface: Additive angular margin loss for deep face recognition.
\newblock In {\em CVPR}, pages 4690--4699, 2019.

\bibitem{DonoserB13Diffusion}
Michael Donoser and Horst Bischof.
\newblock Diffusion processes for retrieval revisited.
\newblock In {\em CVPR}, pages 1320--1327, 2013.

\bibitem{Radenovic-ECCV16}
Radenovi{\'c} Filip, Tolias Giorgos, and Chum Ond{\v{r}}ej.
\newblock {CNN} image retrieval learns from {BoW}: Unsupervised fine-tuning
  with hard examples.
\newblock In {\em ECCV}, pages 3--20, 2016.

\bibitem{Fischler:1981:RSC:358669.358692}
Martin~A. Fischler and Robert~C. Bolles.
\newblock Random sample consensus: A paradigm for model fitting with
  applications to image analysis and automated cartography.
\newblock {\em Commun. ACM}, 24(6):381--395, 1981.

\bibitem{EndToEndDIR2017}
Albert Gordo, Jon Almaz{\'{a}}n, J{\'{e}}r{\^{o}}me Revaud, and Diane Larlus.
\newblock End-to-end learning of deep visual representations for image
  retrieval.
\newblock {\em IJCV}, 124(2):237--254, 2017.

\bibitem{BestPracticeInstanceRetrieval}
Jiedong Hao, Jing Dong, Wei Wang, and Tieniu Tan.
\newblock What is the best practice for cnns applied to visual instance
  retrieval?
\newblock {\em arXiv:1611.01640}, 2016.

\bibitem{resnet}
Kaiming He, Xiangyu Zhang, Shaoqing Ren, and Jian Sun.
\newblock {Deep residual learning for image recognition}.
\newblock In {\em CVPR}, pages 770--778, 2016.

\bibitem{HofferA14TripletLoss2}
Elad Hoffer and Nir Ailon.
\newblock Deep metric learning using triplet network.
\newblock In {\em ICLRW}, 2015.

\bibitem{REMAP}
S.~S. {Husain} and M. {Bober}.
\newblock Remap: Multi-layer entropy-guided pooling of dense cnn features for
  image retrieval.
\newblock {\em TIP}, 28(10):5201--5213, 2019.

\bibitem{bn}
Sergey Ioffe and Christian Szegedy.
\newblock Batch normalization: Accelerating deep network training by reducing
  internal covariate shift.
\newblock In {\em ICML}, pages 448--456, 2015.

\bibitem{IscenATFC18}
Ahmet Iscen, Yannis Avrithis, Giorgos Tolias, Teddy Furon, and Ondrej Chum.
\newblock Fast spectral ranking for similarity search.
\newblock In {\em CVPR}, pages 7632--7641, 2018.

\bibitem{IscenTAFC17}
Ahmet Iscen, Giorgos Tolias, Yannis Avrithis, Teddy Furon, and Ondrej Chum.
\newblock Efficient diffusion on region manifolds: Recovering small objects
  with compact {CNN} representations.
\newblock In {\em CVPR}, pages 926--935, 2017.

\bibitem{JegouDS08Hamming}
Herv{\'{e}} J{\'{e}}gou, Matthijs Douze, and Cordelia Schmid.
\newblock Hamming embedding and weak geometric consistency for large scale
  image search.
\newblock In {\em ECCV}, pages 304--317, 2008.

\bibitem{JegouPDSPS12VLAD}
Herv{\'{e}} J{\'{e}}gou, Florent Perronnin, Matthijs Douze, Jorge
  S{\'{a}}nchez, Patrick P{\'{e}}rez, and Cordelia Schmid.
\newblock Aggregating local image descriptors into compact codes.
\newblock {\em TPAMI}, 34(9):1704--1716, 2012.

\bibitem{KalantidisMO16CroW}
Yannis Kalantidis, Clayton Mellina, and Simon Osindero.
\newblock Cross-dimensional weighting for aggregated deep convolutional
  features.
\newblock In {\em ECCVW}, pages 685--701, 2016.

\bibitem{RMAC}
Zehang Lin, Zhenguo Yang, Feitao Huang, and Junhong Chen.
\newblock Regional maximum activations of convolutions with attention for
  cross-domain beauty and personal care product retrieval.
\newblock In {\em ACMMM}, pages 2073--2077, 2018.

\bibitem{LiuWYLRS17SphereFace}
Weiyang Liu, Yandong Wen, Zhiding Yu, Ming Li, Bhiksha Raj, and Le Song.
\newblock Sphereface: Deep hypersphere embedding for face recognition.
\newblock In {\em CVPR}, pages 6738--6746, 2017.

\bibitem{LiuWYY16SphereFace}
Weiyang Liu, Yandong Wen, Zhiding Yu, and Meng Yang.
\newblock Large-margin softmax loss for convolutional neural networks.
\newblock In {\em ICML}, pages 507--516, 2016.

\bibitem{SGDR2017CosineAnnealing}
Ilya Loshchilov and Frank Hutter.
\newblock {SGDR:} stochastic gradient descent with warm restarts.
\newblock In {\em ICLR}, 2017.

\bibitem{Lowe04SIFT}
David~G. Lowe.
\newblock Distinctive image features from scale-invariant keypoints.
\newblock {\em IJCV}, 60(2):91--110, 2004.

\bibitem{Masi0HN18FaceRecognitionSurvey}
Iacopo Masi, Yue Wu, Tal Hassner, and Prem Natarajan.
\newblock Deep face recognition: {A} survey.
\newblock In {\em SIBGRAPI}, pages 471--478, 2018.

\bibitem{iccvNohASWH17}
Hyeonwoo Noh, Andre Araujo, Jack Sim, Tobias Weyand, and Bohyung Han.
\newblock Large-scale image retrieval with attentive deep local features.
\newblock In {\em ICCV}, pages 3476--3485, 2017.

\bibitem{PerdochCM09HesAff}
Michal Perdoch, Ondrej Chum, and Jiri Matas.
\newblock Efficient representation of local geometry for large scale object
  retrieval.
\newblock In {\em CVPR}, pages 9--16, 2009.

\bibitem{PerronninLSP10FisherVector}
Florent Perronnin, Yan Liu, Jorge S{\'{a}}nchez, and Herv{\'{e}} Poirier.
\newblock Large-scale image retrieval with compressed fisher vectors.
\newblock In {\em CVPR}, pages 3384--3391, 2010.

\bibitem{PhilbinCISZ07Oxford}
James Philbin, Ondrej Chum, Michael Isard, Josef Sivic, and Andrew Zisserman.
\newblock Object retrieval with large vocabularies and fast spatial matching.
\newblock In {\em CVPR}, 2007.

\bibitem{PhilbinCISZ08Paris}
James Philbin, Ondrej Chum, Michael Isard, Josef Sivic, and Andrew Zisserman.
\newblock Lost in quantization: Improving particular object retrieval in large
  scale image databases.
\newblock In {\em CVPR}, 2008.

\bibitem{RITAC18}
Filip Radenovic, Ahmet Iscen, Giorgos Tolias, Yannis Avrithis, and Ondrej Chum.
\newblock Revisiting oxford and paris: Large-scale image retrieval
  benchmarking.
\newblock In {\em CVPR}, pages 5706--5715, 2018.

\bibitem{Radenovic2018FinetuningCI}
Filip Radenovic, Giorgos Tolias, and Ondrej Chum.
\newblock Fine-tuning cnn image retrieval with no human annotation.
\newblock {\em TPAMI}, 2018.

\bibitem{ranjan2017L2constrainedSoftmax}
Rajeev Ranjan, Carlos~D. Castillo, and Rama Chellappa.
\newblock L2-constrained softmax loss for discriminative face verification.
\newblock {\em arXiv:1703.09507}, 2017.

\bibitem{RazavianASC14Offtheshelf}
Ali~Sharif Razavian, Hossein Azizpour, Josephine Sullivan, and Stefan Carlsson.
\newblock {CNN} features off-the-shelf: An astounding baseline for recognition.
\newblock In {\em CVPRW}, pages 512--519, 2014.

\bibitem{LandmarkListWiseLoss}
J{\'{e}}r{\^{o}}me Revaud, Jon Almaz{\'{a}}n, Rafael~Sampaio de Rezende, and
  C{\'{e}}sar~Roberto de Souza.
\newblock Learning with average precision: Training image retrieval with a
  listwise loss.
\newblock In {\em ICCV}, 2019.

\bibitem{FaceNetTripletLoss15}
Florian Schroff, Dmitry Kalenichenko, and James Philbin.
\newblock Facenet: A unified embedding for face recognition and clustering.
\newblock In {\em CVPR}, pages 926--935, 2015.

\bibitem{SivicZ03}
Josef Sivic and Andrew Zisserman.
\newblock Video google: {A} text retrieval approach to object matching in
  videos.
\newblock In {\em ICCV}, pages 1470--1477, 2003.

\bibitem{TeichmannAZS19D2R}
Marvin Teichmann, Andre Araujo, Menglong Zhu, and Jack Sim.
\newblock Detect-to-retrieve: Efficient regional aggregation for image search.
\newblock In {\em CVPR}, pages 5109--5118, 2019.

\bibitem{ToliasAJ16ASMK}
Giorgos Tolias, Yannis Avrithis, and Herv{\'{e}} J{\'{e}}gou.
\newblock Image search with selective match kernels: Aggregation across single
  and multiple images.
\newblock {\em IJCV}, 116(3):247--261, 2016.

\bibitem{ToliasJ14HQE}
Giorgos Tolias and Herv{\'{e}} J{\'{e}}gou.
\newblock Visual query expansion with or without geometry: Refining local
  descriptors by feature aggregation.
\newblock {\em Pattern Recognition}, 47(10):3466--3476, 2014.

\bibitem{WangCLL18CosFaceJournal}
Feng Wang, Jian Cheng, Weiyang Liu, and Haijun Liu.
\newblock Additive margin softmax for face verification.
\newblock {\em SPL}, 25(7):926--930, 2018.

\bibitem{CosFace2018}
Hao Wang, Yitong Wang, Zheng Zhou, Xing Ji, Dihong Gong, Jingchao Zhou, Zhifeng
  Li, and Wei Liu.
\newblock Cosface: Large margin cosine loss for deep face recognition.
\newblock In {\em CVPR}, pages 5265--5274, 2018.

\bibitem{WangSLRWPCW14TripletLoss1}
Jiang Wang, Yang Song, Thomas Leung, Chuck Rosenberg, Jingbin Wang, James
  Philbin, Bo Chen, and Ying Wu.
\newblock Learning fine-grained image similarity with deep ranking.
\newblock In {\em CVPR}, pages 1386--1393, 2014.

\bibitem{Yang2019EfficientIR}
Fan Yang, Ryota Hinami, Yusuke Matsui, Steven Ly, and Shin'ichi Satoh.
\newblock Efficient image retrieval via decoupling diffusion into online and
  offline processing.
\newblock In {\em AAAI}, 2019.

\bibitem{ZhangZQWL19Adacos}
Xiao Zhang, Rui Zhao, Yu Qiao, Xiaogang Wang, and Hongsheng Li.
\newblock Adacos: Adaptively scaling cosine logits for effectively learning
  deep face representations.
\newblock In {\em CVPR}, pages 10823--10832, 2019.

\bibitem{ZhengYT18IRSurvey}
Liang Zheng, Yi Yang, and Qi Tian.
\newblock {SIFT} meets {CNN:} {A} decade survey of instance retrieval.
\newblock {\em TPAMI}, 40(5):1224--1244, 2018.

\bibitem{ZhengZSABBBCN09}
Yantao Zheng, Ming Zhao, Yang Song, Hartwig Adam, Ulrich Buddemeier, Alessandro
  Bissacco, Fernando Brucher, Tat{-}Seng Chua, and Hartmut Neven.
\newblock Tour the world: Building a web-scale landmark recognition engine.
\newblock In {\em CVPR}, pages 1085--1092, 2009.

\end{thebibliography}
}

\end{document}